\documentclass[conference]{IEEEtran}
\IEEEoverridecommandlockouts
\usepackage{cite}
\usepackage{amsmath,amssymb,amsfonts}
\usepackage{algorithmic}
\usepackage{graphicx}
\usepackage{textcomp}
\usepackage[disable]{todonotes}
\usepackage{microtype}
\usepackage{xcolor}
\usepackage{subfigure}
\def\BibTeX{{\rm B\kern-.05em{\sc i\kern-.025em b}\kern-.08em
    T\kern-.1667em\lower.7ex\hbox{E}\kern-.125emX}}
\begin{document}

\title{A Local Approach to Forward Model Learning: \\ Results on the Game of Life Game}


\author{Simon M. Lucas, Alexander Dockhorn, Vanessa Volz, Chris Bamford, Raluca D. Gaina, Ivan Bravi, \\Diego Perez-Liebana, Sanaz Mostaghim, Rudolf Kruse
\thanks{Simon M. Lucas, Vanessa Volz, Chris Bamford, Raluca D. Gaina, Ivan Bravi and Diego Perez-Liebana are with the School of Electrical Engineering and Computer Engineering, Queen Mary University of London, London, UK.}
 \thanks{Alexander Dockhorn, Sanaz Mostaghim and Rudolf Kruse are with the Computational Intelligence Research Group, Otto von Guericke University, Magdeburg, Germany.}
}

\maketitle

\begin{abstract}

This paper investigates the effect of learning a forward model on the performance of a statistical forward planning agent. We transform Conway's Game of Life simulation into a single-player game where the objective can be either to preserve as much life as possible or to extinguish all life as quickly as possible.

In order to learn the forward model of the game, we formulate the problem in a novel way that learns the local cell transition function by creating a set of supervised training data and predicting the next state of each cell in the grid based on its current state and immediate neighbours.  Using this method we are able to harvest sufficient data to learn perfect forward models by observing only a few complete state transitions, using either a look-up table, a decision tree or a neural network. 

In contrast, learning the complete state transition function is a much harder task and
our initial efforts to do this using deep convolutional auto-encoders were less successful. 

We also investigate the effects of imperfect learned models on prediction errors and game-playing performance, and show that even models with significant errors can provide good performance.

\end{abstract}


\section{Introduction}

Learning forward models or world models is a
major challenge for AI that is currently receiving 
significant attention in the research community.

Forward models are immensely powerful tools.
They enable the use of Statistical Forward Planning (SFP) algorithms (such as Monte Carlo Tree Search and Rolling Horizon Evolution) and may be used to generate instant intelligent behaviour
for a wide range of games.  
They can also be used to generate vast quantities of experience data
for reinforcement learning algorithms, which can be combined with
SFP algorithms to generate even stronger intelligence~\cite{Silver2017}.
In addition to the strength of the decision making, forward models 
also lead to more explainable AI, as the likely future system
states can be observed.

Recent work has made significant progress in learning forward models,
more of which will be described in Section~\ref{sec:LitReview}.
Much of this work has used Deep Neural Networks to learn
forward models in the form of entire state transition functions.

However, we take a fundamentally different approach in this paper,
which works well for 2D cellular automata.  The approach involves
two methods that simplify the learning problem:

\begin{itemize}
    \item We learn local update rules rather than the
entire state transition function.  Clearly this is well suited to
the problem at hand, but it provides a relatively simple way to
study the effects of model inaccuracies.  We investigate this
by using models that are trained only on a subset of the possible
local transitions.
\item For these experiments we separate out the simulation model from the player's actions.  This further simplifies the learning problem.
\end{itemize}

Learning the local transition function for 2D cellular automata is
an easy problem due to two facts: the underlying model is a 
good fit with the simulation model, and the transitions are
fully observable.

An important question that we have only started to answer is
how well this approach will work for more complex games,
the next step up being 2D arcade games such as Atari 2600~\cite{bellemare2013arcade} and General Video Game AI (GVGAI)~\cite{perez2018gvgaisurvey} games.

In cases where the forward model is locally learnable, we note that the approach leads to a relative abundance of training data.
Consider the Game of Life (GoL)~\cite{gardner1970a} on an $N \times N$ grid.  The conventional complete state transition learning approach generates 1 training pattern for an encoder architecture per state transition, learning a mapping from an $N \times N$ input to an $N \times N$ output.
Contrast this with the local approach which creates $N \times N$ training patterns per state transition, each one mapping an $L \times L$ input to a single output.  Clearly the local approach vastly simplifies the learning problem: consider a $30 \times 30$ grid for the GoL, which has a $3 \times 3$ local neighbourhood.  When learning the entire state transition function, we have one $900 \mapsto 900$ I/O pair to learn for each observed state transition, compared to $900$ pairs of $9 \mapsto 1$ patterns
for the local case.

A key aspect to this is what the learner is able to observe.
In some games most of the game-state is directly observable.
In other cases, we may have access to the object graph of each
game state (similar to a JSON representation of the game state)
then we may be able to learn good forward models from this,
since in many games the intricate game-play is an emergent
feature of simple object interaction rules.  This is made
explicit in the Video Game Description Language used in GVGAI.

This paper has several novel contributions.  The
first is to formulate the problem of forward model learning
as one of learning a local transition function. This is applied to each element of a state instead of attempting to
learn complete state transition functions.  Secondly,
we show how the player actions can
be decoupled from learning the forward model, 
by applying them independently of the model.  This
further simplifies the learning problem, and can be likened
to learning the laws of physics independently of
the actors in an environment.
Thirdly, in order to
demonstrate the difference this can potentially make
we made a game  by adding player actions and
a scoring function to Conway's Game of Life.  Though
we do not explore it further here, games based on 2D cellular automata may have significant strategic depth
and may perhaps also be fun for people to play.  Finally, we compare the
results of local versus global learning on the Game of Life Game. We also include initial results on further applications of the method on 2 GVGAI games, \textit{Aliens} and \textit{Missile Command}, showing examples where this local learning does \emph{not} work, but suggesting ways in which it might be developed for better performance.


\section{Literature Review}
\label{sec:LitReview}


One of the main benefits of having access to a forward model is the ability to use planning methods such as Monte-Carlo Tree Search (MCTS)
and Rolling Horizon Evolutionary Algorithms.  We use the term Statistical Forward Planning (SFP) to describe such
algorithms due to the way they make decisions based
on the statistics of rollouts (sequences of potential
actions made in a copy of the game state).
These evaluations are then used to choose the 
actions which have high likelihood of increasing the score of the agents.  Compared to conventional reinforcement
learning algorithms, SFP methods have the great benefit 
of providing (in many cases) instant intelligent behaviour
without the need for any learning (Silver~\cite{silver2008sample} refers to MCTS 
as transient learning). 
However, in many environments, the forward model is unknown or it cannot be accessed. In these scenarios the forward model can be approximated or learned.

Deterministic forward models in some cases can be learned by tabular methods \cite{attias_2003} \cite{sutton_1991}.
These methods learn the exact state transition functions by building a look-up table, similar to the \emph{Exact Learner} we use here (though a key aspect of this paper is to apply the look-up table learning method at a local rather than a global level).  Tabular methods are very accurate, but are limited to problem domains that have relatively small and discrete state and action spaces when applied globally.

To solve the problem of estimating the state transition function of high dimensional or continuous state spaces, several rule-based heuristic methods have been proposed.

In \cite{dockhorn_2018}, Hierarchical Knowledge Bases are used to build a rule-based representation of the several games in the GVGAI framework. These estimated forward models are then used in conjunction with MCTS and Breadth First Search in order to plan the next best moves. 
Even when only modelling a small set of features, learned models were able to improve the agent's performance despite the limited accuracy of predicting the next game state.
Models with higher prediction accuracy were based on  ensembles of decision trees, but made use of pre-filtered feature sets \cite{Dockhorn2018}.
Similarly in \cite{uriarte_2015}, rule-based methods are used to create a forward model of combat models in StarCraft. These rule based methods have the benefit of learning very compact representations, sometimes only consisting of a few hundred rules to encompass the dynamics of a single game. Due to the minimal representations, they also can have very fast inference times. However, these methods also require knowledge of the environment in order to engineer features such as sprite types and unit properties.

Deep Neural Networks (DNN) can also be used to predict game states simply by observing the pixels of frames and learning the state transition model to the next frame. DNN models of environments tend to contain the following components: an encoder that encodes the state into a latent variable space, a decoder that takes the latent variable space and outputs the next state, a way of encapsulating the action or actions from a single time frame, and a recurrent component that can track time-dependent internal variables. This architecture is used in \cite{oh_2015} and \cite{chiappa_2017} in order to predict the next frames of several Atari games in the Arcade Learning Environment (ALE). Both of these methods are able to predict hundreds of states into the future, but eventually the simulated state of the game diverges from the actual state.
This means that using simulation methods such as MCTS that require roll-outs have to limit roll-out length, as larger lengths are less accurate. This issue can be detrimental in games where rewards are sparse and any rewards could only be observed in very long roll-outs. 
An attempt to decouple time from individual time-steps is proposed in \cite{gregor_2018}, where a novel temporal difference auto-encoder is used to predict environmental states at arbitrary time-steps. 


A particularly relevant use case for forward models is modelling planning through imagination. Imagination in this context is the process of learning an internal model of the environment it is in and then using that model to plan trajectories or interpret the results of many possible trajectories. In \cite{weber_2017} and \cite{buesing_2018} models of the environments are initially generated using supervised learning. These models are then used to generate possible new trajectories, the trajectories are then interpreted using another neural network feeding into the policy of the agent. 
\cite{weber_2017} Also shows that agents with access to models without perfect accuracy are still able to perform better than agents using algorithms such as MCTS.


\section{Making a Game from the Game of Life}

Despite the name, Conway's Game of Life (GoL) is not actually a game in the
conventional sense since there are no player actions.  
The GoL simulation has been studied from many viewpoints; although several applications involving player actions exist\footnote{One example of GoL allowing user input: http://chromacon.surge.sh}, the impact of user interaction has not been thoroughly studied, as far as we know. Additionally, even versions which allow for user input do not include reward functions or win/loss identification.

GoL involves a 2D grid of cellular automata where the state of each
cell at time $t+1$ depends only on its current state and the state
of its eight immediate neighbours.  There are many possible update
rules that involve only this neighbourhood.  The cell states are binary,
so given the 9 cells there are $2^9 = 512$ possible local patterns.  Every possible
update rule can therefore be expressed as a truth table with $512$ rows,
and each output (next cell state) can be either 0 or 1.  Hence there are
$2^{512}$ possible update rules, though some of these are equivalent to 
each other under reflection, rotation and inversion.  GoL is therefore
one highly studied update rule in a large space of possible update rules.

In this paper we consider both GoL and also a rule that has
recently been used for Procedural Content Generation (PCG),
in particular to generate cave-like structures~\cite{johnson2010cellular} - we'll
call this rule CG for Cave Generator.

The GoL update rule is presented in Equation~\ref{eq:gol}, where $X$ is the current cell being updated and $N$ is the current number of neighbours of this cell which are alive.

\begin{equation}
    X =
    \begin{cases}
      1, & \text{if}\ X=1~and~(N=2~or~N=3)  \\
      1, & \text{if}\ X=0~and~(N=3)\\
      0, & \text{otherwise}
    \end{cases}\label{eq:gol}
\end{equation}

Equation~\ref{eq:cave} presents the CG update rule, where $X$ is the current cell being updated, $N$ is the current number of neighbours of this cell which are alive, and $T$ is a threshold (here, as in~\cite{johnson2010cellular}, $T$=4).

\begin{equation}
    X =
    \begin{cases}
      1, & \text{if}\ N>T  \\
      0, & \text{otherwise}
    \end{cases}\label{eq:cave}
\end{equation}

\begin{figure}[!t]
\centering
\includegraphics[width=0.9\columnwidth]{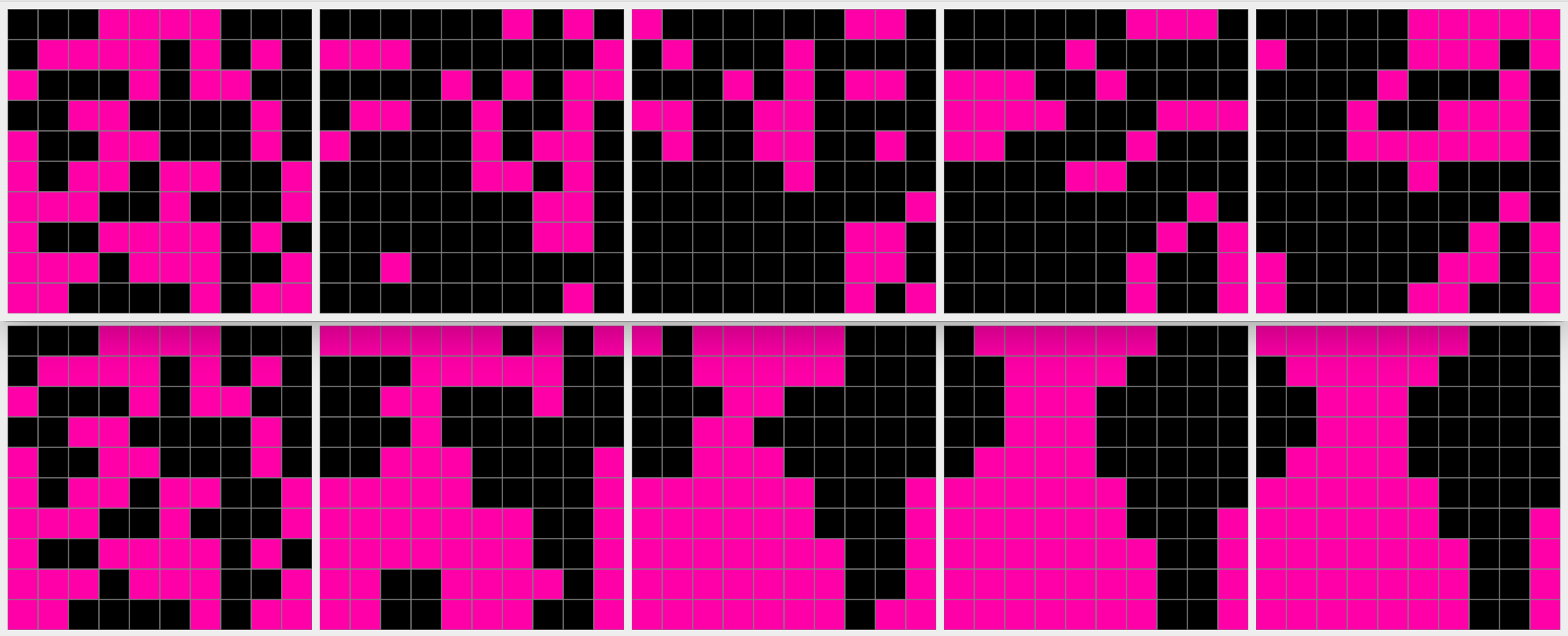}
\caption{\label{fig:CaveAndGoL}
 Sample game state trajectories for GoL (top) and CG (bottom).  
}
\end{figure}

We select these two rules because they have some inherent interest
and because they differ significantly in their complexity and
how hard they are to learn for a classifier.  See Figure~\ref{fig:CaveAndGoL} for example trajectories for the 2 distinct update rules.

We can transform GoL into a game by adding player actions and a reward structure. At each time step a player may do nothing, or select any cell on the grid to flip the state of that cell (so a 0 becomes a 1 and vice versa). The simulation is then updated as before. At each time step, the game score is calculated by counting the number of cells which are alive (have a value of 1).  The game is then to either nurture as much life as possible by aiming to maximize the score, or to eliminate all life and hence achieve a minimum score of zero.
We can apply this to any 2D cellular automata simulation.

\section{Methods}

\subsection{Evaluation}

There are two distinct methods under test: the conventional
complete state transition function learner, and the local learner.

We consider three evaluation methods:  

\begin{enumerate}
    \item \textbf{Supervised training}.  From a set of random start grids, run the
    system forward using the true model to observe a set of state transitions.
    This gives a set of input/output training pairs. The complete (global) state learner gets as input just one large vector pair per state transition, while the 
    local learner obtains $N \times N$ per transition.  The local learner has a single output to predict, while the complete learner has to predict all $N \times N$ outputs.  In each
    case, we report the average hamming distance between the target output and the learned model's output.
    
    \item \textbf{N-step prediction test}.  Once a model has been trained, we can then run it forward, again from a set of random start states, and predict $T$ steps into the future.  Here we used $T=30$, which we observed to be long enough to distinguish perfect models from near-perfect ones. This depends on the update rule, with highly non-linear rules such as GoL often diverging quickly (it depends on the initial state and on which truth table entries differ).
    
    \item \textbf{Game playing performance}.  We play the game using a Rolling Horizon Evolutionary Algorithm (RHEA) from a set of $100$ random start grids and play the game for $100$ steps.  The RHEA agent uses the learned model instead of the true one.
\end{enumerate}

From the perspective of developing strong game AI, the final evaluation is the most important, as it focuses on how well the learned model works with the SFP agent. As we'll see, we can also tune agents to cope better with erroneous models.
Furthermore, we know from other cases that models can be woefully 
inaccurate and still lead to good game play performance in an SFP agent.  The key insight
is that if a model's errors do not change the rank order of the estimated 
action values, then they have no effect on the decisions made.

\begin{figure}[!t]
    \centering
    \includegraphics[width=\linewidth, trim={0 0 0 2cm},clip]{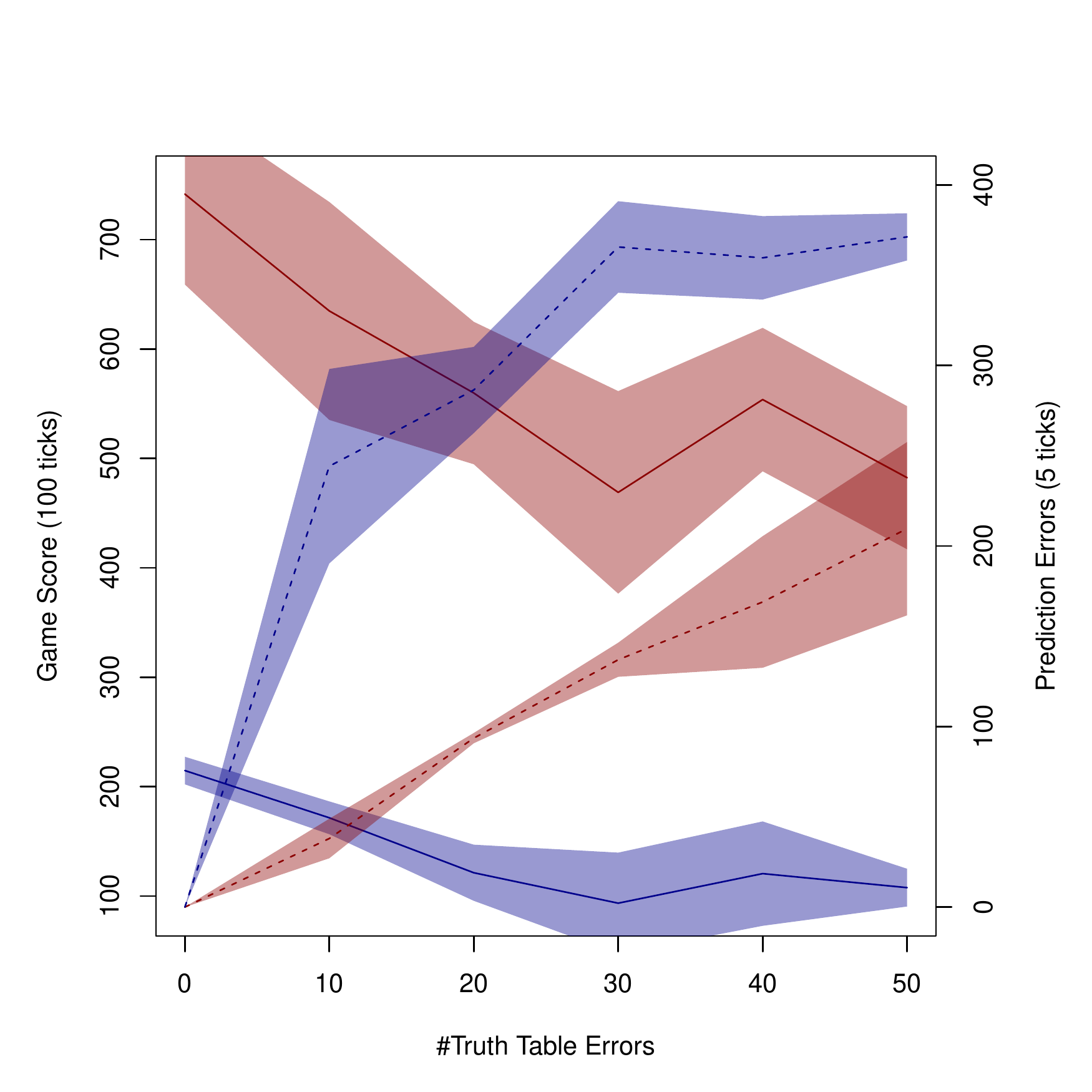}
    \caption{Number of truth table errors plotted against game score after 100 ticks (solid lines) and against the number of prediction errors after 5 ticks (dashed lines). Results using the Game of Life update rule are depicted in blue, while red corresponds to the Cave Generator update rule.}
    \label{fig:truth_error}
\end{figure}

In the case of 2D cellular automata, the local learning approach enables some
more efficient ways to estimate performance.  Since the model
can be completely described using a truth-table with just $512$ 
entries, we can compute the truth table for the learned model
(irrespective of the nature of the learned model e.g. decision
tree, random forest, support vector machine or neural network).
This enables us to compile the learned model into a truth table,
which we then use  in the SFP agent with identical results.  The
advantage of doing this is two-fold.  Firstly, it is much faster.
Secondly, we can analyze the relationship between truth table
differences and performance differences. See Figure~\ref{fig:truth_error} for a visualisation of truth table errors plotted against the game score and number of prediction errors, for both GoL and CG update rules.

In the rest of this paper, we focus on the GoL update rule as it contains more complex dynamics than the CG update rule, though in future it would be interesting to compare how different update rules affects the ability of learning algorithms to generalise from small samples of the local transition data.

\subsection{Game Playing Agent}

For the game-playing agent, we used a Rolling Horizon Evolution agent based on a $1+1$~EA.
This is a simple and fast agent that has proven to be effective
across a range of games. It was used in \cite{lucas2019efficient} to compare the performance of model-based hyper-parameter optimization algorithms on the game \textit{Planet Wars}. In \cite{lucas2018n}, this RHEA agent was tuned using the N-Tuple Bandit Evolutionary Algorithm (NTBEA) for two games: \textit{Asteroids} and \textit{Planet Wars}. A more complex population-based RHEA agent was shown to achieve high performance on range of GVGAI games as well~\cite{gaina2017rolling}.
The version we use here is extended from~\cite{lucas2019efficient}. It evolves a single action sequence by repeatedly mutating it. The original is replaced with the mutated sequence if, after simulating through the sequence, the heuristic value of the final state reached is better than the original solution. One action sequence may be evaluated $nEvals$ times, to reduce noise. A shift buffer may be used to keep the final solution evolved in one tick to the next (instead of starting from a new random action sequence), by removing the first action in the sequence and adding a new random action at the end, in order to keep a constant length. They key extension in this work is the addition of a mutation transducer.  When creating a mutated copy
of a sequence, this operator can either copy an action from the parent sequence, insert
a random action (with $probMutation$ probability), or copy the previous action of the new sequence (with $repeatProb$ probability).  This provides a soft
form of macro-actions. During mutation, the algorithm may be forced to mutate at least one action in the sequence. All of these settings and all parameter possible values are presented in Table~\ref{tab:space}.
\todo[inline]{Explain discountFactor parameters.}

\begin{table} [!t]
\centering
\caption{\label{tab:space}RHEA Hyper-Parameters.}
\begin{tabular}{cccp{2.5cm}}
\hline
ID & Parameter & Type & Legal values \\
\hline
$\phi_0$ & $flipMinOneValue$ & boolean & $false, true$ \\
$\phi_1$ & $probMutation$ & double & 0.1, 0.2, 0.3, 0.4, 0.5, 0.6, 0.7 \\
$\phi_2$ & $sequenceLength$ & integer & 1, 3, 5, 10, 20 \\
$\phi_3$ & $nEvals$ & integer & 1, 3, 5, 10, 25\\
$\phi_4$ & $shiftBuffer$ & boolean & $false, true$ \\
$\phi_5$ & $mutationTransducer$ & boolean & $false, true$ \\
$\phi_6$ & $repeatProb$ & double & 0.1, 0.2, 0.3, 0.4, 0.5, 0.6, 0.7\\
$\phi_7$ & $discountFactor$ & double & 0.999, 0.99, 0.9, 0.8\\
\hline
\end{tabular}
\end{table} 

Since the algorithm should be evaluated at the best of its capabilities, we used NTBEA to tune RHEA under two circumstances: using a model with all 512 patterns (perfect model) and one with only 480 (near perfect model).  For these experiments we used the exact learner (see below), noting that we can obtain the equivalent results for other learning algorithms applied to the locally sampled data by compiling them into truth tables.

For each condition we run the algorithm 100 times. For each execution, NTBEA was given a budget of 100 evaluations to search the space of 28800 configurations. The fitness of a configuration is given by the number of cells alive at the end of a single game. NTBEA was configured with $k=300$ and $\epsilon=0.5$.
\todo[inline]{Simon, Ivan: compare the model learned from 480 patterns
with the correct truth table.} 



\subsection{Local Learning Methods}

\subsubsection{Exact Learner}

The exact learner memorizes the output for each set
of inputs. If it has not seen a pattern before, it
returns the \emph{a priori} most likely output, which for GoL is $0$.

Using this type of learner provides insight into
how the performance of the agent varies with respect to model
inaccuracy, as we know the model will be perfectly accurate
for all the patterns it has seen, and will output a default
value for all unseen patterns.
This learner does not assume any regularity in the structure of
the function to be learned.  In this sense it provides a good
contrasting baseline for other classifiers that do attempt
to generalize.

\subsubsection{Decision Tree}
A decision tree classifier \cite{Berthold2010} is a fast and reliable classifier that should be able to replicate the game of life patterns reasonably well.
While playing the game, every observed pattern is added to the training set of the decision tree.
Since the decision tree is trained to completely fit the observed training examples (no pruning), it is able to learn the environment's dynamics after several game ticks.

Due to the fast learning time, the agent builds a new decision tree every time a new game state transition was observed (adding $N\times N$ observations to the training set, though
the number of unique patterns (512) quickly saturates for the current experiments).
Experiments have shown that the built model is able to quickly generalize from the seen examples, which lets the agent predict previously unseen patterns and
learn a perfect model.

\subsubsection{Neural Network}


Initial experiments were made using a neural network
with a single hidden layer to learn the local patterns.
It was also able to learn a perfect model.
More work is needed to see whether neural
networks are able to learn in a more sample-efficient
way than decision trees or other supervised
learning methods when applied to the local patterns.




\subsection{Complete State Transition Function Learning}

An alternative to local learning is to learn the
complete state transition function using a Deep Convolutional Auto-Encoder (DCAE).
Learning complete state transition functions
is the more conventional approach within forward model learning.

So far we have only made some preliminary experiments,
but clearly learning a complete state transition
function is much harder than learning local transitions
in cases where the local transitions can accurately
capture the true model.

Several experiments were run on grids of size 10x10 and 20x20. The auto-encoder was able to reconstruct the state with high accuracy from a latent state space of 256 and 512 neurons respectively after $100,000$ state transition observations.   However this architecture was sample inefficient and unstable to train.  Difficulty of training  increased significantly with respect to the grid size, meaning that a 30x30 grid could only be trained to low accuracy given our experimental setup. The reason for this difficulty is likely to be that an auto-encoder network tries to generalize across the entire image space and may be trying to memorize all the state transitions instead of learning the local rules. The linear latent layer in state-space models such as this would have to be able to accurately encode and correctly transition all $2^{900}$ possible states of the grid. 
Further work would be needed to make the training stable enough to function reliably on larger grid sizes such as a 30x30 grid.






\begin{figure}[!t]
\centering
\includegraphics[width=0.9\linewidth]{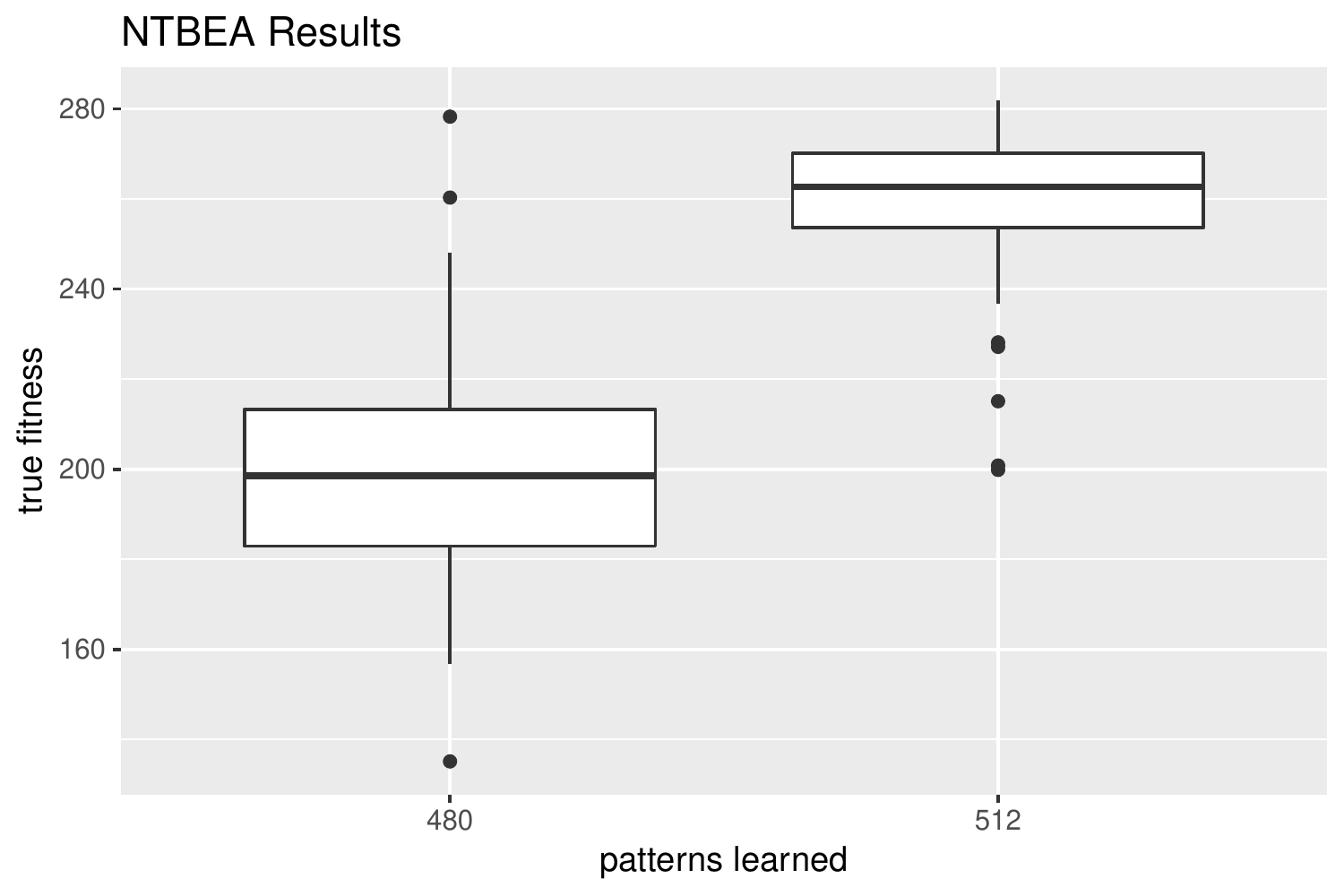}
\caption{\label{fig:ntbea}
Results of hyper-parameter tuning of the RHEA agent with two different amounts of learned patterns. The box plot shows the true fitness distribution of the configuration suggested by 100 independent NTBEA runs.}
\end{figure}

\begin{figure}[!t]
\centering
\includegraphics[width=0.9\linewidth]{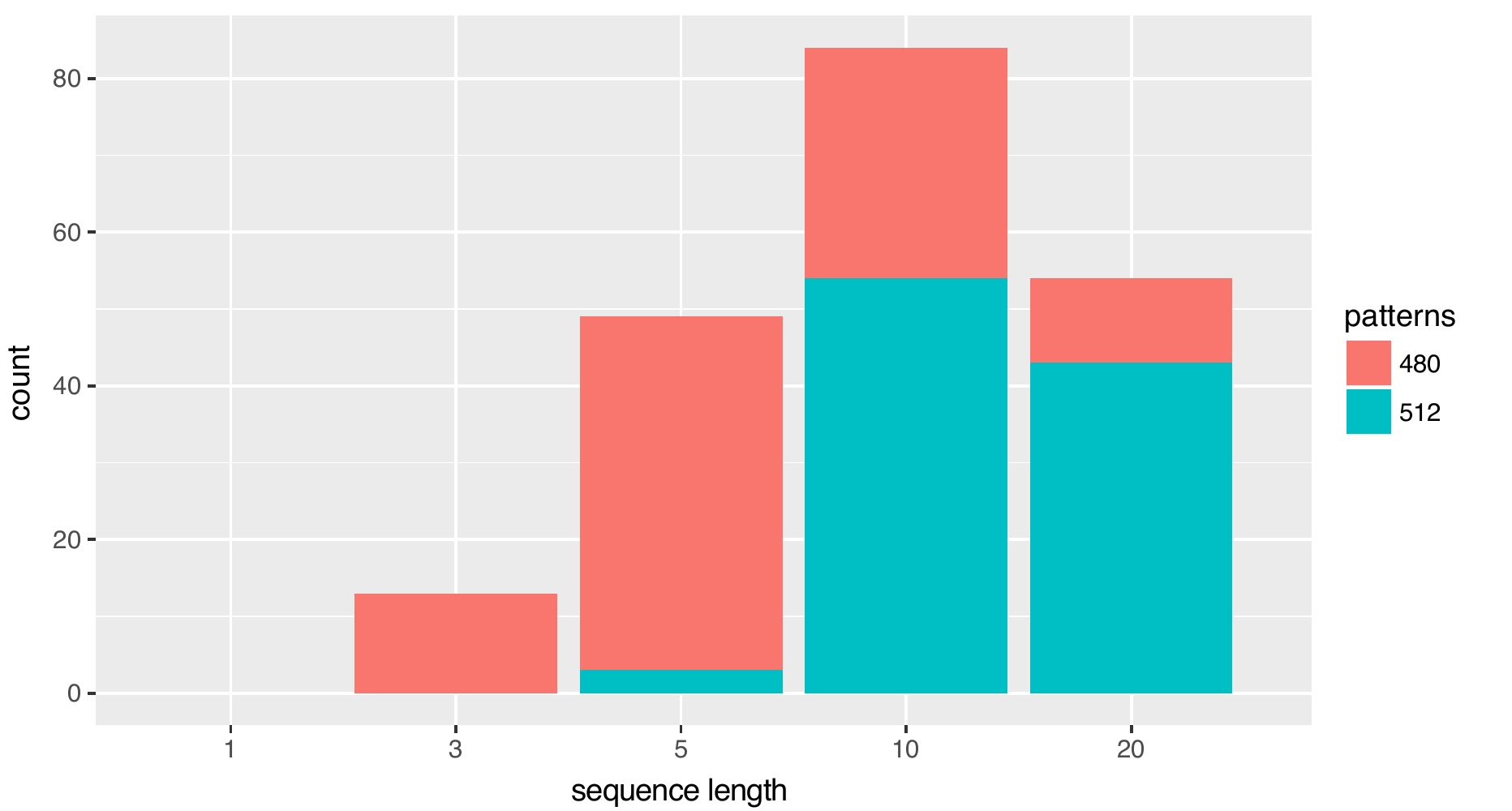}
\caption{\label{fig:ntbea_length} Distribution of the sequence length parameter values when using 480~(red) or 512~(green) patterns learned.}
\end{figure}

\section{Results}

The outcome of the hyper-parameter tuning experiments is shown in Figure~\ref{fig:ntbea}. We can clearly see how providing a less accurate forward model significantly degrades RHEA's performance even when its parameters are tuned. 
Mining the configurations returned for the two conditions, we could highlight an interesting feature: when the forward model is imprecise, it is better to evolve shorter sequences in order to reduce the noise in RHEA's fitness signal. In fact, the prediction error accumulates as the length increases, causing a more noisy evaluation of action sequences. 
In Figure~\ref{fig:ntbea_length}, the two distributions can be observed as skewed towards shorter and longer lengths, respectively, for the 480 and 512 case. 
The final configuration used for the rest of the experiments is the best resultant from tuning with a perfect model: $\phi_0$=$true$, $\phi_1$=0.3, $\phi_2$=20, $\phi_3$=25, $\phi_4$=$true$, $\phi_5$=$false$, $\phi_6$=0.2, $\phi_7$=0.8.



\begin{figure*}[!t]
\centering
\includegraphics[page=1, width=0.4\textwidth]{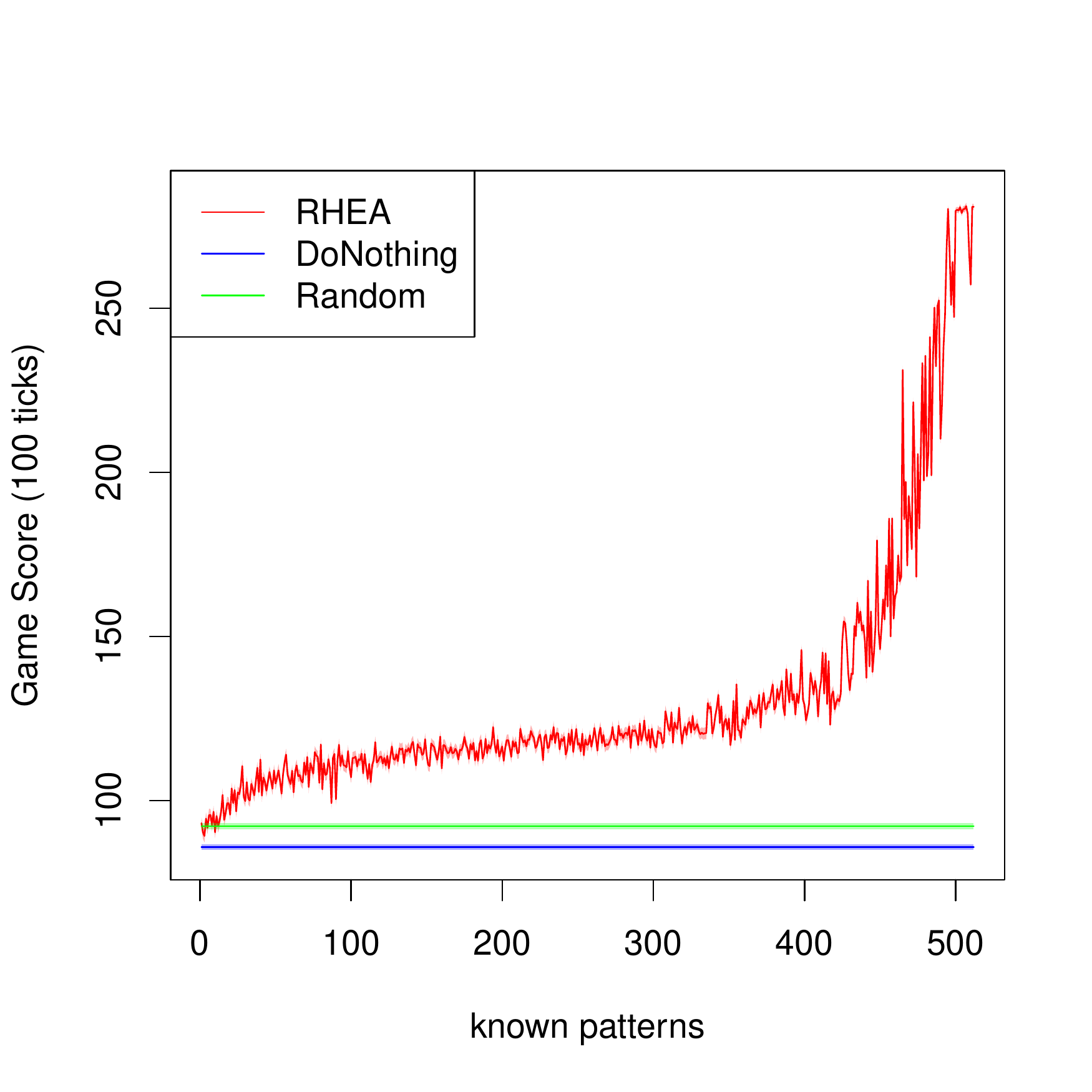} \vspace{1cm}
\includegraphics[page=3, width=0.4\textwidth]{Pics/lutSizePlots.pdf} \\[-1.5cm]
\includegraphics[page=2, width=0.4\textwidth]{Pics/lutSizePlots.pdf} \vspace{1cm}
\includegraphics[page=4, width=0.4\textwidth]{Pics/lutSizePlots.pdf}\\[-1cm]
    \caption{Visualisations of game score (top row) and number of prediction errors (bottom row) with relation to the number of known patterns (left column) and game tick (right column).}
    \label{fig:lutSize}
\end{figure*}

Figure \ref{fig:lutSize} shows the relation of the achievable agent performance and the number of known patterns. 
For each model trained, 15 games were simulated for 100 game ticks. This process was then repeated 50 times, the results averaged per game tick. All lines include a visualisation of $\pm 1.5$ of the standard deviation in a lighter colour (which is mostly low and thus only faintly visible). The plots also include baseline comparisons with the DoNothing and Random agents.

For the chosen RHEA configuration, the number of prediction errors decreases approximately linearly with the increase in known patterns, except for a slight increase when very few patterns are known (see Figure~\ref{fig:lutSize} bottom left). \todo{shouldn't this always increase? Might be due to the fact that which patterns are chosen is pretty random. But then the sd should be higher?} In contrast, we see a slow increase in game score until approximately 80\% of possible patterns are known, whereafter the agent performance increases rapidly (see Figure~\ref{fig:lutSize} top left).

RHEA performance is consistently better than that of the baseline agents for the whole runtime of the game (see figure \ref{fig:lutSize} top right). However, the plot also shows that before the rapid performance increase after 80\% of patterns are known, all agents lose points until they are able to stabilise. This stabilisation could also be helped by the fact that the number of distinct patterns seems to decrease with runtime of less proficient players, thus also reducing the number of prediction errors (see figure \ref{fig:lutSize} bottom right).

\begin{figure}[!t]
\centering
\includegraphics[width=0.9\columnwidth]{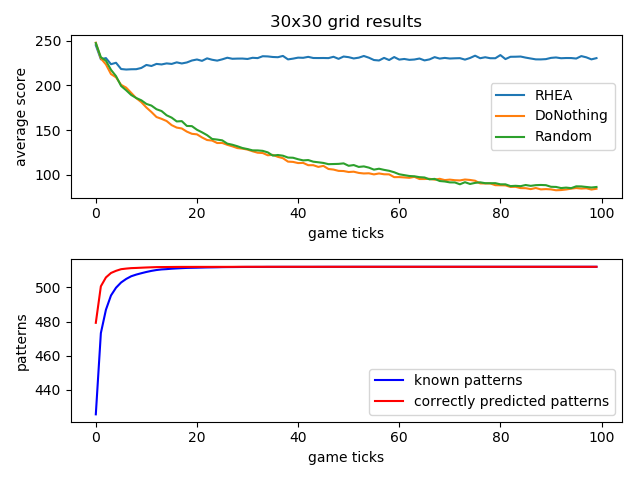}
\caption{\label{fig:LearnerComparison}
(top) Performance comparison of a RHEA agent using a decision tree model learned while playing the game, an agent behaving randomly, and an agent that does nothing.
(bottom) Average number of observed and correctly predicted patterns by the decision tree.
The left hand part of the graph shows that the correctly predicted patterns grows much more quickly than the
number of observed patterns, showing that the decision
tree is generalising to unseen patterns.
}
\end{figure}

In contrast to tuning a model before playing the game, we analyzed the agent's playing performance while training a new model at every game tick.
Figure~\ref{fig:LearnerComparison} compares the performance of a RHEA agent using a Decision Tree model with a random agent and an agent that does nothing.
As the comparison shows, the agent using a Decision Tree quickly learns to predict the environment and can effectively keep the population alive.

\section{Further Applications}

The approach presented in this paper has shown a very high proficiency in the GoL game, as the section above shows. It is well suited to the local forward model approximation used in the proposed algorithm, due to the dynamics of the game from state to state. Therefore, it is worth questioning how well this approach scales to more complex and diverse games. 

An appropriate benchmark to test the generality of this approach would be the General Video Game AI (GVGAI) framework, which includes more than $150$ games of various types~\cite{perez2018gvgaisurvey}. Compared to the GoL game, GVGAI games are significantly more complex, although most interactions happen at a local level as well. Some global interactions do exist (such as the presence of resources that can be picked up to use later), which would not allow a strictly local pattern learning method to succeed. One example could be \textit{Race Bet}, a game in which a race happens between certain sprites; the avatar, placed far away from the race, needs to step on a tile with the matching colour of the sprite which is going to win. Interactions in this case are not local: one racer reaches a goal on one point of the screen, while the avatar must move in another area. Other games also pose problems of partial observability, such as \textit{Kill Bill Vol 1} and \textit{Eighth Passenger}, which may introduce noise into the system.

There are, however, quite a few games with local patterns for which we believe this approach can provide good results. We have run a preliminary study in two GVGAI games: \textit{Aliens} and \textit{Missile Command}. The former is a stochastic game, a variant of Space Invaders, in which the player controls a ship moving at the bottom of the screen (action set: left, right and shoot) aiming to kill all the enemies before they reach the ship. \textit{Missile Command}, a port of the game with the same name, is a deterministic game where the player controls a ship which is free to move in all 4 directions and create an explosion in the direction the avatar is facing. Several missiles drop from the top of the screen in the direction of several cities on the ground. The player needs to prevent all cities from being hit by destroying the missiles before they reach them. The player wins if at least one city is saved, and loses if all cities are gone or if the ship itself gets hit by a missile.

One thing to note is that, even if games with local interactions only are used, the process of encoding the game rules, terminations, scoring and pattern validation is not trivial. Ideally, a learning system should learn not only transitions between states, but also reward functions, win/loss identification methods, illegal situations and effects of the avatar actions (currently actions are manually inserted into a game state before learning the state transitions $S \mapsto S'$).



Both games are fairly similar in concept and difficulty for a vanilla RHEA agent (100\% win-rate in both, first level only considered for \textit{Missile Command})~\cite{gaina2017analysis}. However, their complexity in terms of local patterns does differ. In Aliens, there are 7 possible values a cell in a pattern might take, if we consider all sprites relevant for interactions (empty, avatar, missile, alien bomb, protective base, alien or end of screen), which leads to over \textit{40 million} ($7^9$) possible patterns. Many of these are not valid within the game's rules, however: for example, there cannot be more than 1 avatar in a pattern. Such simple counting rules can reduce the space to just over \textit{2 million}. On the other hand, there is an upper bound on the number that can be observed of $(T \times W \times H)$ given $T$ state transitions of a game grid of dimension $(W \times H)$.
 
\begin{figure}[!t]
\centering
\subfigure[Aliens]{\includegraphics[width=0.49\columnwidth]{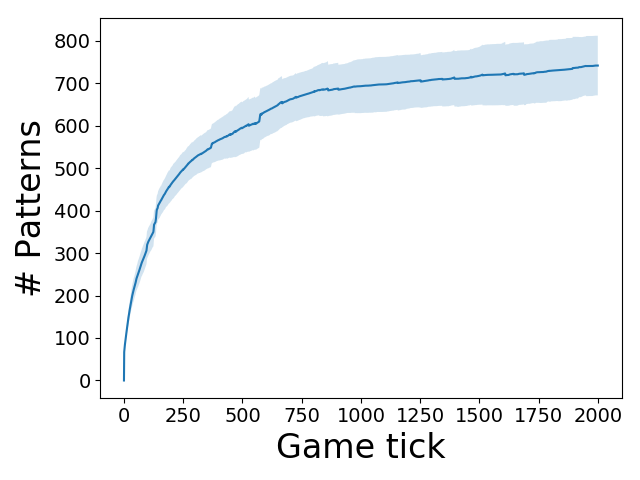}\label{fig:aliens-patterns}}
\subfigure[Missile Command]{\includegraphics[width=0.49\columnwidth]{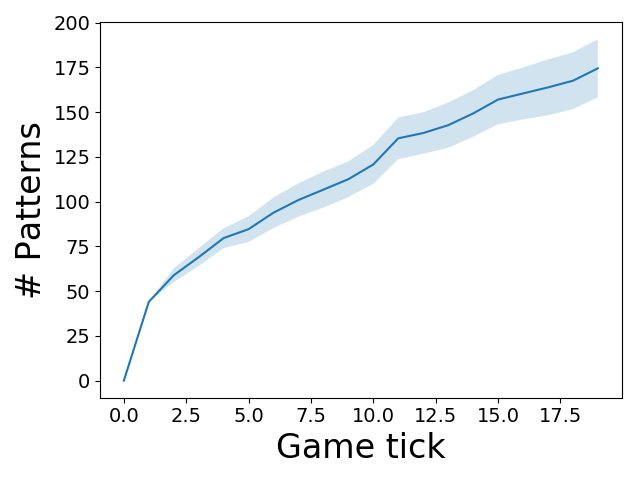}\label{fig:mc-patterns}}
\caption{Average number of patterns discovered per game tick by a random player, over 100 runs of a game. The shaded area indicates standard deviation.
}
\end{figure}

In our experiments, agents can observe a maximum of 2300 patterns (summed across 1000 games played).
Figure~\ref{fig:aliens-patterns} shows the average number of patterns discovered per tick during 100 games. It is interesting to observe the shape of the curve, with several patterns being discovered quickly, while the progress slows down about a third into the game. 



In Missile Command there are 6 sprites to be considered in a pattern (empty, wall, city, missile, explosion or avatar), for a total of \textit{10 million} total patterns. These can be further reduced in a similar way as before, with simple sprite counting rules, to \textit{4 million}. However, agents discover only a maximum of 2800 patterns in 1000 games played.
Figure~\ref{fig:mc-patterns} shows the average number of patterns discovered per tick during 100 games. Although the total number is much lower than in Aliens, so is the average game length, which gives the agent less data points to learn from; the curve also appears it would continue to grow if given more game time, different to the plateau observed in Aliens. 

Current results using a RHEA agent with a learned model on these two games are not satisfactory, showing a player strength at or below random. 
The main complexity in modelling the \textit{Aliens} game rules comes from the uncertainty about the movement of the aliens. These could be moving either left or right, but this information is not captured in a local 3x3 pattern. This could be solved by encoding additional information, stacking frames, or increasing the pattern size, to reduce the negative impact of incorrect simulations. A further complexity imposed by Missile Command is the legal movement of the avatar outside of the screen: the player's ship may leave the game area and return later, which would signal a game loss to game rules coded based on the grid observed by the agent playing the game.


An immediate point for future work is to investigate how learning local patterns can be enhanced to accommodate for more complex environments like these GVGAI games. Our findings suggest that more efficient generalization techniques should be used when predicting unknown transitions (rather than assuming that nothing changes). Although the agents do not discover many of the possible patterns when playing GVGAI games, it's possible that those that \textit{are} discovered contain key rules which can be applied in similar, even if not identical, scenarios. A possible way forward would be to use \textit{attention} techniques~\cite{mnih2014recurrent} to learn patterns over 2 or more convolutional kernels, each one of them centered in different and relevant locations of the game.

\section{Conclusions and Future Work}

We have seen two extreme examples of performance of the proposed local learning approach: success in the Game of Life Game, and initial failure on two GVGAI games.  In the Game of Life Game,
the local model is a perfect fit for the true model of the game.  The Exact Learner (a simple approach which simply stores all patterns observed and their corresponding transition, with no generalisation) was able to learn a perfect forward model after observing only 20 state transitions.  The Decision Tree method was able
to learn perfect local forward models even more quickly.

It was further able to learn a model accurate enough for good performance when used within a Rolling Horizon Evolutionary Algorithm (RHEA) after observing a single state transition: much better than random and around 80\% of performance given a perfect forward model. Similarly, a RHEA agent using a Decision Tree learner can very quickly correctly predict the environment and greatly outperform a random player. Thus, a powerful feature of local transition function learning is that we only need to observe a few simulation ticks in order to sample the majority (perhaps even all) of the possible patterns.


In contrast, our initial attempts to learn
a complete state transition function for $30 \times 30$
grids were not successful here, even
after training on $100,000$ state transitions.
Learning the complete state transition function
is clearly a much harder problem.  On the other
hand, while learning local models is much easier
when it is possible, there may be scope for
hybrid approaches that capture aspects of local
learning while going beyond it only when necessary.


We presented preliminary results of a direct
and rather simplistic application of the method
(using only 3x3 local patterns in the current frame and not accounting for further game complexities)
to two GVGAI games, \textit{Aliens} and \textit{Missile Command}. This led to models that
were pathologically bad, showing worse
than random performance when used within a Rolling Horizon Evolution agent instead of the game's true forward model.  However, on closer
inspection, this need not be too concerning:
we believe that local models can still work
well for these games, providing they are given
enough context in which to make good predictions.
This context would be provided in the form of
an extended feature vector: for example, providing
a larger than 3x3 window or using frame stacking to be able to determine 
sprite directions.

It will be interesting to investigate whether these
features can be made in a general way for a wide
class of 2D games or need to be engineered or
evolved for each game. Future work will thus look at learning more than simple state transitions and incorporate reward functions, pattern validation and action effects in the knowledge learned by the system, for a complete and general learner.

Finally, when using an SFP agent such
as RHEA with any forward model,
be it imperfect or perfect, the parameters
should be tuned to optimise performance
not only for the problem at hand but also
for the learned forward model.  For example,
we noted that smaller values for the sequence length parameter
were consistently selected by the NTBEA optimiser when using imperfect forward models.

\section*{Acknowledgment}

This work was partially funded by the EPSRC CDT in Intelligent Games and Game Intelligence (IGGI) EP/L015846/1.


\bibliographystyle{IEEEtran}  
\bibliography{references}

\end{document}